\documentclass[twocolumn,showpacs,prl]{revtex4-1}
\usepackage{amsmath}
\usepackage{amsfonts}
\usepackage{graphicx}

\usepackage{color}
\usepackage{algorithm2e}
\usepackage{comment}

\begin{document}
\title{Permutation-equivariant neural networks applied to dynamics prediction}
\author{Nicholas Guttenberg}
\affiliation{Araya, Tokyo}
\affiliation{Earth-Life-Science Institute, Tokyo}
\author{Nathaniel Virgo}
\author{Olaf Witkowski}
\affiliation{Earth-Life-Science Institute, Tokyo}
\author{Hidetoshi Aoki}
\author{Ryota Kanai}
\affiliation{Araya, Tokyo}

\begin{abstract}
The introduction of convolutional layers greatly advanced the performance of neural networks on image tasks due to innately capturing a way of encoding and learning translation-invariant operations, matching one of the underlying symmetries of the image domain. In comparison, there are a number of problems in which there are a number of different inputs which are all 'of the same type' --- multiple particles, multiple agents, multiple stock prices, etc. The corresponding symmetry to this is permutation symmetry, in that the algorithm should not depend on the specific ordering of the input data. We discuss a permutation-invariant neural network layer in analogy to convolutional layers, and show the ability of this architecture to learn to predict the motion of a variable number of interacting hard discs in 2D. In the same way that convolutional layers can generalize to different image sizes, the permutation layer we describe generalizes to different numbers of objects.
\end{abstract}

\maketitle 

\section{Introduction}

\begin{figure}[t]
\includegraphics[width=\columnwidth]{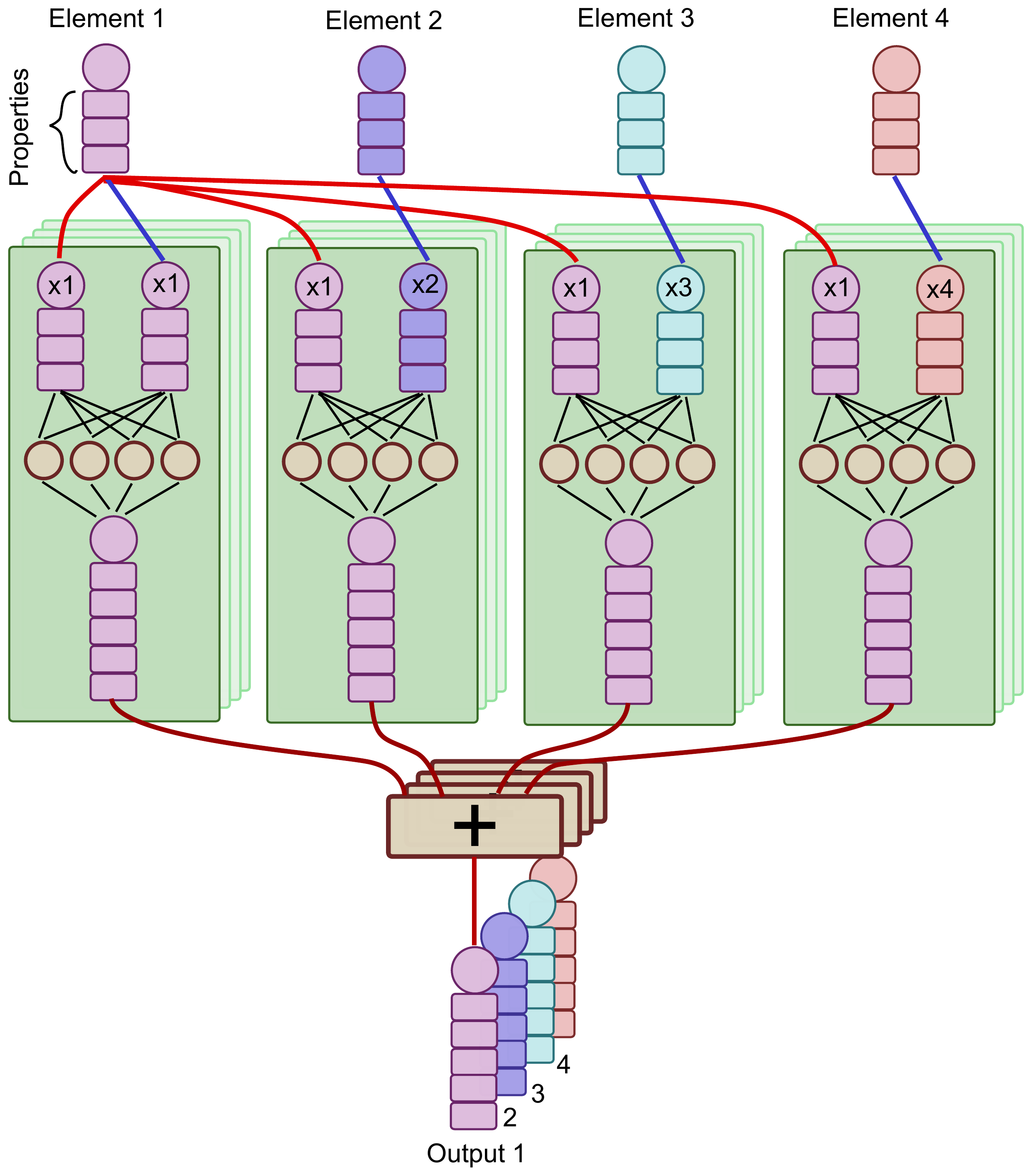}
\caption{\label{PLayer}Structure of a permutational layer. Each output $y_i$ is calculated with a function that separately combines the correspondingly indexed input $x_i$ with each of the other inputs $x_j$, and then is summed over $j$. The same weights are used for all $i,j$ combinations. }
\end{figure}

Problems with a high degree of symmetry create large inefficiencies for machine learning algorithms which do not take those symmetries into account. When the problem has symmetries (such as translation or rotation invariance, or in the case of this paper, permutation invariance), then the algorithm must independently learn the same pattern multiple times over, leading to a massive increase in the amount of training data and time required, and creating many opportunities for spurious biases to creep in. One approach is to handle this at the level of the data, removing the redundant symmetries from the input and aligning things into a canonical form. Examples of this range from things like data augmentation\cite{scholkopf1996incorporating}, SIFT features\cite{lowe1999object} and spatial transformer networks\cite{jaderberg2015spatial}. Attentional models also do this implicitly, by aligning the location of attention to features rather than coordinates in the input data\cite{graves2014neural,graves2016hybrid,bahdanau2014neural,luong2015effective,merity2016pointer}.

The other possibility is to build the desired symmetry into the functional form of the network at some level. For example, convolutional neural networks do this with respect to translation invariance. Because the same shared filter is applied at each point in the image space, and then (eventually) aggregated through a hierarchy of pooling operations, the final classification result behaves in a locally translation-invariant way at each scale. This kind of convolutional approach has been extended to other domains and other symmetries. For example, graph convolutional networks allow for convolution operations to be performed on local neighborhoods of arbitrary graphs\cite{defferrard2016convolutional,kipf2016semi} (this kind of approach can be used for both invariance and, see later, equivariance). A general framework for constructing these kinds of specified invariances was proposed in \cite{gens2014deep}, where the authors generalized convolutional operations to affine invariant computations and showed a way to construct similar things for arbitrary symmetry spaces. For permutation symmetry, work along this line has been done in specifying an expansion in terms of a neural net applied to specific symmetry functions for estimation of potentials in ab initio quantum mechanics calculations\cite{jeong2015data}. In addition, \cite{edwards2016towards} uses pooling over instances to construct permutation-invariant summary statistics for different data distributions.

Instead of invariance, which generates a result that does not change under transformations of the input, one might want to obtain a property called `equivariance'. In this case, the goal is not to make the final outcome constant despite the application of some set of transformations, but rather to create a structure such that the input and output of the network both behave the same way under transformation. That is to say, if invariance targets a function of the form $f(x) = f(T[x])$, equivariance targets a function of the form $f(T[x]) = T[f(x)]$. This is useful for things such as image segmentation, where applying a rotation to the input image should not change the pixel classes, but should change the positions at which those classes occur (in the same way as the input was rotated). A general recipe for constructing equivariant convolutions was proposed in \cite{cohen2016group}, where this technique was used to generate a convolutional layer equivariant to 90 degree rotations and to mirror symmetry. 

In this paper, we're interested in making networks which predict the future dynamics of sets of similar, interacting objects --- for example, learning to predict and synthesize the future dynamics of a set of particles just by observing their trajectories. This could be something along the line of deriving effective equations of motions for complex particles such as animals in a swarm or players of a sport, or apply at a more abstract level to something such as relationships between variations in different sectors of the economy or in the interaction between specific stocks. The associated symmetry to this sort of data is permutation symmetry, and we are looking for methods which let us build permutation equivariant neural networks.

We present numerical experiments assessing the feasibility and performance of various approaches to this problem, and show that a combination of multi-layer embedded functions inside a max pooling permutation invariant wrapper can achieve significantly higher accuracy on predicting the dynamics of interacting hard discs. Furthermore, networks using this combination demonstrate the ability to generalize to a different number of particles than seen during training, and can also handle cases in which the particles are not identical by way of including an auxiliary random label feature.

\section{Permutational Layers}

The common theme to constructing group-invariant or group-equivariant functions is to perform a pooling over all combinations of the symmetry transforms. In the case of a symmetry group, applying a transformation associated with that symmetry only shuffles around the members of the group --- it does not add or remove a member. That means that as long as one does a calculation that is applied identically to all members of the symmetry group, the output will be invariant to transformations applied to the input. Equivariance can be constructed in a similar fashion by combining two members of the input space and only doing this pooling with respect to transformations applied to one of the two.

For permutation equivariance, there is some existing work, which differ in how interactions between elements are considered. Variadic networks\cite{mcgregor2007neural,mcgregor2008further} implement permutation equivariance through the addition of a kind of mean-field approximation of the interactions between elements to otherwise single-element neural networks (but the mean-field effect extends to hidden variables as well, so may indirectly capture higher order interactions). Recent work on set convolutions\cite{ravanbakhsh2016deep} also provides an equivariant layer for interchangeable objects, in which summation or maximization over pairs of objects is used to construct the equivariant functions.

Both of these approaches have in common that there is a kind of wrapper which guarantees that the influence of the rest of the elements on a particular member of the set is pooled in an invariant way, but generally the structure outside of the wrapper is made complex and the structure within the wrapper is made simple (being either an unstructured mean over the population or a maximization or summation of linear functions of the population features). 

When considering interacting particles, this may run into difficulties --- interaction potentials may be, relatively speaking, fairly complex functions. But once we apply a pooling operation, we will lose the ability to track exactly what particles were interacted with in order to contribute to a given hidden layer feature. In that sense, it might require an excessive number of layers to reconstruct the correct interactions, because the network has to learn how to use the hidden variables to encode not just the interaction but also to store some kind of indexing information over its neighbors (similar to the way that pooling forces an autoencoder in the image domain to store positional information inside the learned features so that it can be reconstructed later during the decoding process). 

The ability to wrap a function with a pooling operation such that the output is invariant is quite general though, so it should be possible to put a complex function inside the interaction --- a multi-layer neural network. This is somewhat analogous to approaches in the image domain which apply 'convolutions' with a filter size of 1 (sometimes referred to as network-in-network) in order to increase the complexity of operations done at a particular scale, for example in the Inception architecture\cite{szegedy2015going}. There have also been recent experiments showing that factoring convolutions between the spatial part and local part can be advantageous\cite{chollet2016xception}.

\begin{figure}[t]
\includegraphics[width=\columnwidth]{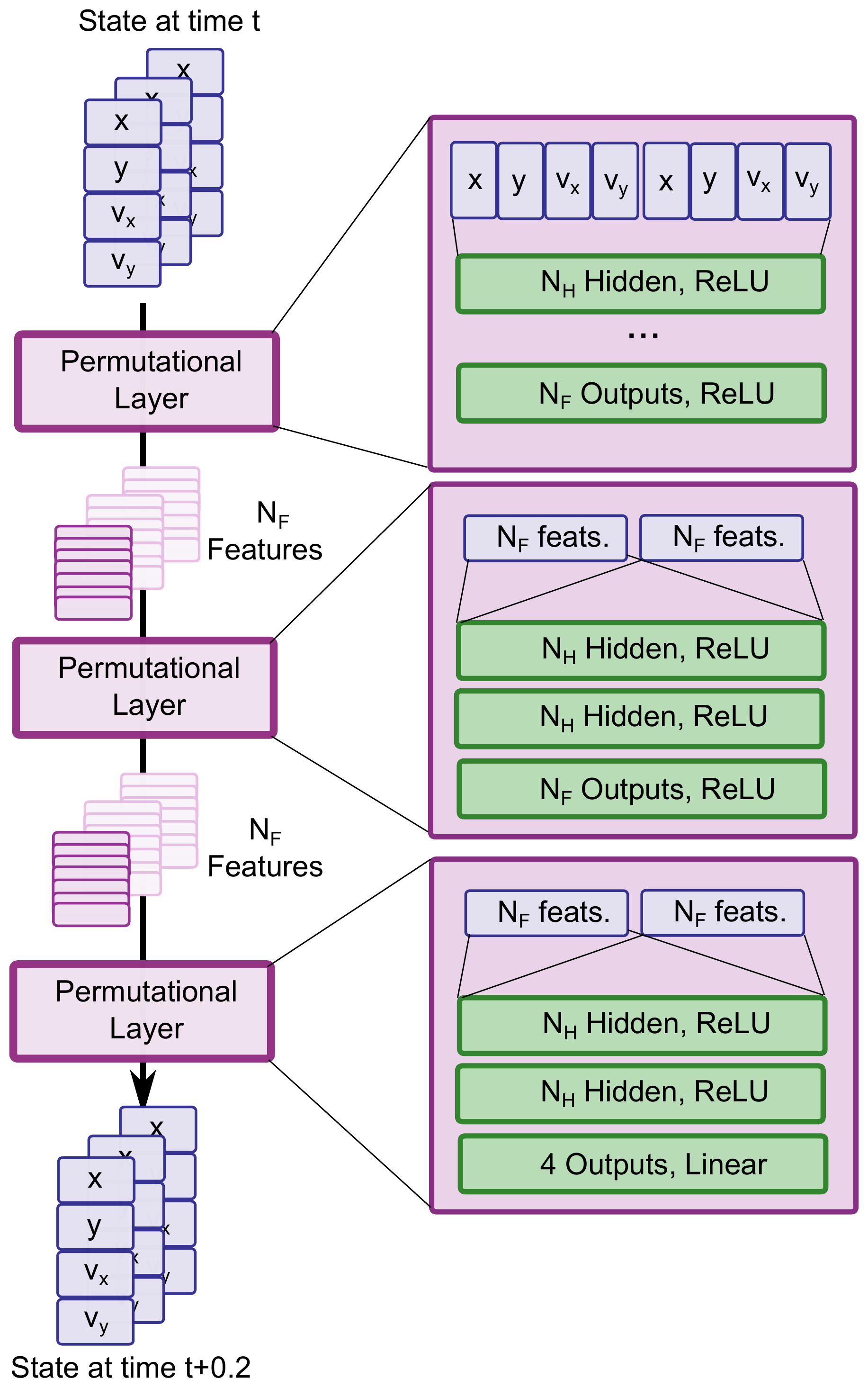}
\caption{\label{NetworkEx}Example network architecture involving permutational layers. Inputs to the permutational layer are a rank-2 tensor containing $N_O$ objects and $N_1$ input features. Within a permutational layer is a neural network with $2N_1$ inputs, and $N_2$ outputs, which is run on features taken from all $N_O^2$ pairs of objects. The result is $N_2$ output features on all $N_O$ objects. These layers can be stacked to capture multi-object interactions.}
\end{figure}

We construct a permutation equivariant layer in a similar way as the form described in \cite{ravanbakhsh2016deep} in the sense that pairs of elements are chosen from the input set, and these results are aggregated over all indices of the second element. However, in their work, they apply matrix multiplications to both input elements and then sum the result to create the object which is aggregated to form the output of the layer (their Eq.~3, paraphrased here):

$$ \vec{y_i} = \sigma( \sum_j \mathrm{W_1} \vec{x_j} + \mathrm{W_2} \vec{x_i}) $$

where $\mathrm{W_1}$ and $\mathrm{W_2}$ are matrices of parameters, $\sigma$ is an element-wise nonlinearity, and the summation over $j$ can be replaced with anything with the appropriate properties over the set of elements (for example, maximization).

In our case, we instead stack the two input elements together and then apply a dense neural network, then pool over the output of that network applied to all choices of the second element. That is to say, a single layer takes all combinations of two input elements $\vec{x_i}$ and $\vec{x_j}$, and generates an output:

\begin{equation}
\vec{y_i} = \frac{1}{N} \sum_j f(\vec{x_i}, \vec{x_j})
\label{PermutationalLayer} 
\end{equation}

where $f$ is an arbitrary function, e.g. a multi-layered dense neural network. We will also consider a version where rather than a sum, we use: $y_i = \max_j f(x_i, x_j)$.

The use of permutation invariance as a wrapper around an arbitrarily complex function is the central idea of this paper; from here, we'll refer to this as a 'permutational layer' (Fig.~\ref{PLayer}). These layers can be stacked to form a more complex network (Fig.~\ref{NetworkEx}).

Interestingly, because this is only in terms of pairwise selections from the input set, it is only necessary to consider $O(N^2)$ interactions overall. On the other hand, the method of \cite{gens2014deep} would at first glance have an $N!$ cost due to having to construct the full symmetry space associated with all permutations of the input set. What are we missing by looking at only $N^2$ interactions?

We note that there is a similar situation in traditional convolutional neural networks. The most general convolutional layer would have filters of the same size as the input image, which would require require $O(N^4)$ computations to evaluate. However, by restricting the filters to be fixed size and local in nature, the cost is reduced back to $O(N^2)$. What we're doing here is somewhat similar --- the full symmetry space is constructed by taking all combinations of pairwise exchanges (sort of like taking into account all vectors which can separate pairs of pixels in the full convolution). But we can consider only a finite number of pairwise exchanges as a sort of local neighborhood. In this case, viewed from the point of view of the method of \cite{gens2014deep}, the pairwise layer is equivalent to a sort of nearest-neighbor filter on the symmetry space.

In the image domain, this would restrict the receptive field quite strongly, as stacking multiple layers at the same resolution just causes the filter sizes to add. But in the case of permutation symmetry, all elements are densely connected, and so information percolates through the set by pairwise interactions very quickly. The output of a 2-body interaction such as Eq.~\ref{PermutationalLayer} generates latent features which could be thought of as containing information about pairs of particles. However, the next layer can effectively read that information for different pairs, meaning that each output hidden feature after two layers can contain at most information about the interaction between sets of 4 particles. This upper bound continues to grow exponentially with network depth. Of course, the information in the hidden features is not likely to be as simple as information about a single specific other particle but is rather a kind of population average weighted in a way which depends on the features of the receiving particle. Because the network can learn to construct that population average in a way that is relevant to each receiving particle separately, it becomes much easier to express locally relevant summary statistics rather than a common mean field.

\begin{figure}
\includegraphics[width=\columnwidth]{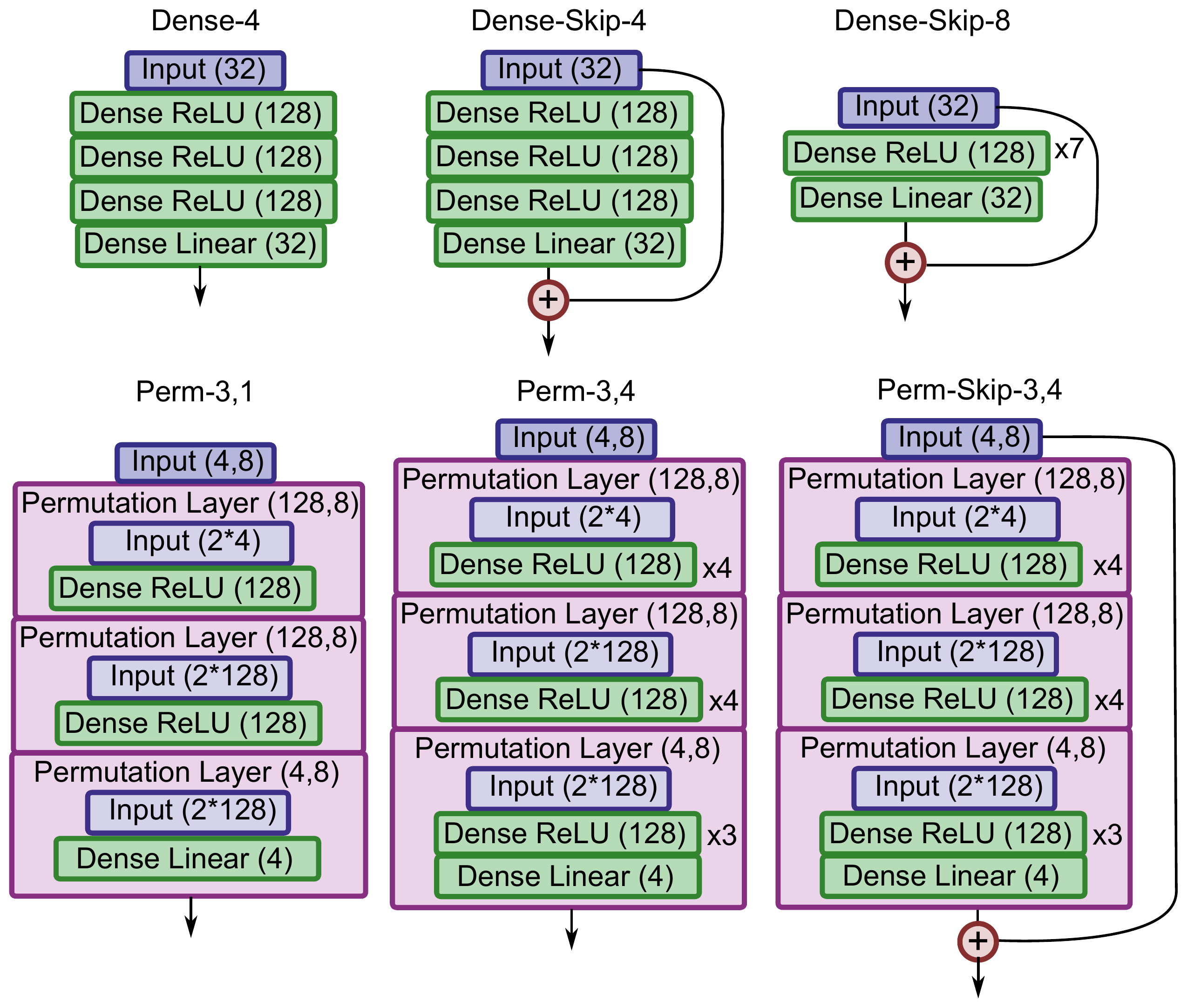}
\caption{\label{NetworkArchs}Network architectures we investigate for the hard disc dynamics task. We compare dense networks of different sizes with and without a final skip connection, as well as a 3-layer permutational network with and without a final skip connection. We also consider a version of Perm-Skip-3,4 with max pooling permutational layers instead of average pooling.}
\end{figure}

\section{Hard disc dynamics}

To test this architecture, we build a network to predict the dynamics of a set of $N$ colliding hard discs in 2D. We simulate the dynamics using PyODE (\url{http://pyode.sourceforge.net}, \url{http://ode.org}), using a variable number of discs of radius $0.2$ in a box $[-1,1]$ with coefficient of restitution $0.9$, zero friction, and a timestep of $0.02$. We initialize the system, simulate for $200$ steps to relax the positions of the discs, and then set the initial velocities to $(\mathcal{N}(0,1), \mathcal{N}(0,1))$, followed by $400$ steps of simulation to generate a training trajectory. The task of the network is to take the positions and velocities of all of the discs at one point in time as input, and to predict the positions and velocities $10$ timesteps later (using mean squared error for comparison).

We compare the performance of various network architectures on this task, each trained on $20000$ trajectories of length $400$. The networks are implemented in Lasagne\cite{lasagne} and Theano\cite{theano}. We train using the Adam optimizer\cite{kingma2014adam}, with a learning rate schedule $r = 10^{-3} \exp(-t/5000)$, where $t$ is the number of training steps. In all cases, inputs to the network are $x,y,v_x,v_y$ for each of the $8$ discs, and outputs are the predicted values at $t+0.2$. We compare three dense architectures (two using a residual skip connection\cite{he2015deep,long2015fully} to effectively calculate only the changes in coordinates, one directly predicting the future coordinates) and four architectures using permutational layers. The network architectures we examine are shown in Fig.~\ref{NetworkArchs}. 

In the case of the permutational layer networks, we use a stack of three such layers in each case. However, we compare the results for when the layers contain only a single matrix multiplication followed by a ReLU nonlinearity, versus cases in which the layers contain 4-layer dense neural networks. We also compare the difference between using average pooling to aggregate over all the pairwise combinations, versus using max pooling. 

In general we found that for this kind of task, increasing the complexity of a single permutational layer works better than adding additional permutational layers, suggesting that capturing the precise form of the inter-particle interactions is more important to accurate prediction than capturing complex multi-body interactions. Furthermore, max pooling appears to both perform better and to generalize significantly better than average pooling.

\begin{figure}
\includegraphics[width=\columnwidth]{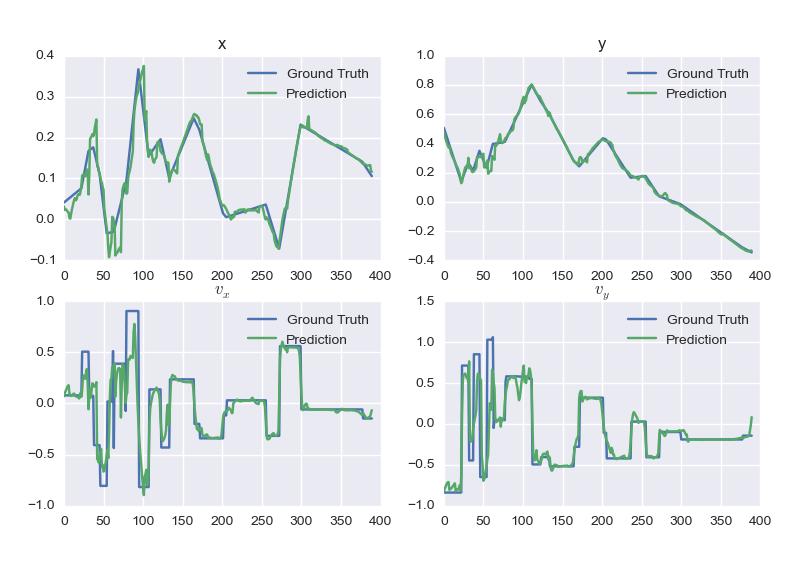}
 \caption{\label{Trajectories}Example predicted (10 steps) and actual trajectories, versus time (timesteps) for the Perm-skip-3,4 architecture on an 8 disc system. }
\end{figure}

\begin{table}
 \begin{tabular}{|cccc|}
\hline
Network & 4-disc & 8-disc & 12-disc \\
\hline
Dense-4 & 0.069 & 0.095 & 0.073\\
Dense-Skip-4 & 0.067 & 0.088 & 0.071\\
Dense-Skip-8 & 0.066 & 0.092 & 0.072\\
Perm-3,1 & 0.047 & 0.062 & 0.063 \\
Perm-3,4 & 0.014 & 0.028 & 0.036\\
Perm-Skip-3,4 & 0.013 & 0.025 & 0.038\\
Perm-Skip-3,4-Max & \textbf{0.011} & \textbf{0.018} & \textbf{0.035} \\
\hline
\end{tabular}
\caption{\label{ResultTable}Final errors of the different network architectures on the hard-disc task for different numbers of discs, averaged over the last 10000 epochs of training.}
\end{table}

Our training results are shown in Table \ref{ResultTable}. All methods appear to more or less converge at the same rate, but the dense architectures perform significantly worse on the task. Furthermore, increasing the depth of the dense networks beyond a certain point makes the asymptotic convergence worse. In both cases, using a skip connection from the beginning of the calculation to the end improves performance. Direct inspection of predicted trajectories show that the velocities are harder to predict than positions, and that the moments of collision with the walls and with other discs are the difficult points to resolve accurately. Predicted trajectories versus actual trajectories are shown in Fig.~\ref{Trajectories}. 

We can also use network to generate a fictitious trajectory by taking its predictions as the next input state to generate the overall sequence. Accuracy compared to the real trajectory quickly decays, but overall the resultant configurations are qualitatively consistent with the idea that these objects are hard discs --- overlaps are avoided, the discs reflect off of one another and appear to transfer momentum to eachother in doing so (in that motion tends to spread from a point of collision). It appears as though the model has captured at least some qualitative statistical regularities of the motion of hard discs, sufficient to generalize to extended trajectories. An example of a synthesized trajectory generated using the Perm-Skip-3,4-Max architecture can be seen at \url{https://youtu.be/s2DwiwrR1qU}.

\textbf{Generalization}:~ Because the permutational networks are defined as an operation which applies uniformly to all inputs, much like convolution they are flexible to the size of the input (the number of particles in this case). However, this is not a guarantee of generalization, just a statement that the action of the network is still well-defined for any number of input objects. We test this property by taking networks trained on between $4$ and $12$ discs and using them to predict the motion of different numbers of discs. We measure the mean squared error of the Perm-Skip-3,4 and Perm-Skip-3,4-Max networks' predictions when using it to predict a different number of discs than it was trained on, averaged over 1000 unseen sequences. The results are shown in Table \ref{Generalization}. For numbers of objects within a factor of 2 of the training case, the network appears to retain some predictive ability (comparable to the error of competing dense network architectures trained directly on that case). Outside of that range, the network's performance rapidly degrades.

The way in which the network's generalization ability fails may relate to the choice of pooling used in defining the permutational layer structure. We use average pooling in order to aggregate the different interactions, but other types of pooling would also be permutation invariant and would extract different types of statistics from the population of objects (for example, max pooling is used in the affine symmetry network in \cite{gens2014deep}). Average pooling conceals information about the absolute number of particles present compared to an un-normalized sum, while max pooling might focus on something closer to nearest-neighbor interactions. Here we restrict ourselves to only average pooling as our focus is to present the permutational layer, but these other forms of pooling would be reasonable architectural choices to try in order to further optimize generalization performance. Varying the number of objects during training is another possible strategy.

\begin{table}
\begin{tabular}{|c|ccccc|}
\multicolumn{6}{c}{Average pooling} \\
\hline
 & 2-disc & 4-disc & 6-disc & 8-disc & 12-disc \\
\hline
4-disc & 0.046 & 0.011 & 0.037 & 0.065 & 0.147 \\ 
8-disc & 0.154 & 0.091 & 0.031 & 0.024 & 0.052 \\
12-disc & 0.370 & 0.147 & 0.083 & 0.052 & 0.036 \\
\hline
\end{tabular}

\begin{tabular}{|c|ccccc|}
\multicolumn{6}{c}{Max pooling} \\
\hline
 & 2-disc & 4-disc & 6-disc & 8-disc & 12-disc \\
\hline
4-disc & 0.0036 & 0.010 & 0.018 & 0.026 & 0.037  \\ 
8-disc & 0.0046 & 0.0078 & 0.012 & 0.017 & 0.027 \\
12-disc & 0.012 & 0.018 & 0.024 & 0.028 & 0.031 \\
\hline
\end{tabular}
\caption{\label{Generalization}Generalization of the Perm-Skip-3,4 network to different numbers of discs. The row corresponds to the training case, the column to the test case. Entries are the average mean-squared-error over 1000 unseen sequences of dynamics. The max pooling version appears to have significantly better generalization ability to different numbers of discs (and surprisingly actually works better than models trained on the same number of discs in some cases). In this system, the true dynamics are strongly determined by nearest-neighbor interactions, which may contribute to the usefulness of max pooling compared to average pooling.}
\end{table}

\textbf{Weak permutation invariance}:~ Another consideration of this kind of problem is that objects are not always identical. That is to say, it may be that for the most part objects or data elements can be swapped, but there are some latent features which make the objects behave somewhat differently. These features may be known, or it may be that they must be learned from data. If the features must be learned but are persistent for a given object from example to example, then a permutation invariant architecture cannot directly capture that (much like a convolutional architecture can't directly use different weight values for different positions in the image).

However, this functionality can be added back in while preserving generalization over the permutation invariant parts of the system. This can be done by assigning additional features to the input data to act as object labels. This could be as simple as a unique vector associated with each object, that the network must learn to associate with the latent properties, or it could be more advanced. For example, one could use a low-dimensional embedding learned from the individual object trajectories using, e.g., PCA, tSNE\cite{maaten2008visualizing}, or a siamese network\cite{bromley1993signature}. 

To test this idea, we train Perm-skip-3,4 on a system composed of $8$ discs of radius $0.1$ and $4$ discs of radius $0.2$ (all having the same mass). We compare the results when the network only has $x,y,v_x,v_y$ versus when we also provide each disc an arbitrary 2-component random vector label. The base network obtains an average mean-squared-error after training of $0.041$, significantly worse than the identical discs case. On the other hand, the augmented network obtains a MSE of $0.022$ which is about the same performance as the un-augmented algorithm achieves on identical discs. So even a very simple sort of auxiliary labeling can allow non-permutation-invariant information to be injected back into the permutation invariant architecture while retaining benefits in places where permutation invariance is appropriate. Of course, with this sort of random label it is unlikely that the network would generalize to different numbers of discs with new random labels, but that choice was made just to test if the network could learn these heterogeneities on its own. One could for example directly label discs with their radius instead, and that should generalize (similarly, a low-dimensional embedding that clusters different kinds of objects to similar points in the feature space could generalize to different sets of objects, so long as their coordinates in that feature space were known ahead of prediction).

\section{Conclusions}

Permutation equivariant network architectures have been used for a variety of cases where the inputs and outputs are a set of interchangeable objects. We considered a problem of this type, in learning to predict the trajectories of sets of interacting objects. 

We note that there are two sources of complexity that the network must learn to handle. One is complexity arising from choices of complicated subsets of elements for multi-element interactions. The other is complexity arising from the form of the interaction itself. For the problem of dynamics prediction, the interactions are often captured well by a pairwise approximation but the specific form of the interactions can be a complicated function of the properties of the particles. Because of this, simple linear functions of pairs of particles must struggle to capture something such as the precise position of the collision between two discs.

To handle this, we observed that the idea of taking a direct sum over pairwise selection of elements from a set to create permutation equivariance can act as a general wrapper around any sort of function combining those elements --- it does not have to be a linear function or sum of the elements with eachother. We can use this wrapper to embed a deep neural network in order to learn and model the form of complicated interaction potentials. We find as a result that this architecture can capture complex interactions more accurately, and in a way which readily generalizes to different numbers of non-identical objects.

For future investigation, it would be interesting to examine other problems in the domain of permutation invariance and closely related symmetries (such as in the case of graph convolutions), to see whether the complexity of the pairwise interactions or the complexity of the multi-body relationships is the dominant factor in performance for those problems.

The code for the network architectures and layers described in this paper along with the numerical experiments on hard disc dynamics and a trained Perm-Skip-3,4-Max network are available at \url{https://github.com/arayabrain/PermutationalNetworks}.

\section{Acknowledgements}

We would like to acknowledge Thomas Kipf and Michaël Defferrard for helpful comments on the content and presentation of this paper.

\bibliography{references}

\end{document}